\documentclass[11pt,a4paper]{article}
\usepackage{acl2015}
\usepackage{times}
\usepackage{url}
\usepackage{latexsym}
\usepackage{multirow}
\usepackage{amssymb,amsmath,graphicx,subfigure,amsfonts,epsfig,epstopdf,datetime}
\usepackage{float}
\usepackage{color}
\title{Relation Classification via Recurrent Neural Network}

\author{Dongxu Zhang$^{1,3}$, Dong Wang$^{*1,2}$ \\
  $^1$CSLT, RIIT, Tsinghua University   \\
  $^2$Tsinghua National Lab for Information Science and Technology\\
  $^3$PRIS, Beijing University of Posts and Telecommunications \\
  {\tt {wangdong}99@mails.tsinghua.edu.cn} \\}

\date{}

\begin{document}
\maketitle
\begin{abstract}
  Deep learning has gained much success in sentence-level relation classification.
  For example, convolutional neural networks (CNN) have delivered competitive performance
  without much effort on feature engineering as the conventional pattern-based methods.
  Thus a lot of works have been produced based on CNN structures. However, a key issue
  that has not been well addressed by the CNN-based method is the lack of capability
  to learn temporal features, especially long-distance dependency between nominal pairs.
  In this paper, we propose a simple framework based on recurrent neural networks (RNN)
  and compare it with CNN-based model. To show the limitation of popular used SemEval-2010
  Task 8 dataset, we introduce another dataset refined from MIML-RE\cite{2014emnlp-kbpactivelearning}.
  %To enhance the model, we present several modifications
  %including a max-pooling approach and a bi-directional architecture.
  Experiments on two different datasets strongly indicates
  that the RNN-based model can deliver better performance on relation classification, and it
  is particularly capable of learning long-distance relation patterns. This makes it
  suitable for real-world applications where complicated expressions are often involved.
\end{abstract}

\section{Introduction}

This paper focuses on the task of sentence-level relation classification.
Given a sentence $X$ which contains a pair of nominals $ \langle x,y \rangle $,
the goal of the task is to predict relation $r \in R$ between the two nominals $x$ and $y$,
where $R$ is a set of pre-defined relations~\cite{hendrickx2009semeval}.

Conventional relation classification methods are mostly based on pattern matching,
and an obvious disadvantage is that high-level features such as tags of part of speech (POS),
name entities and dependency path are often involved.
These high-level features require extra NLP modules that not only increase computational cost,
but introduce additional errors. Also, manually designing patterns is always time-consuming
with low coverage.

Recently, deep learning has made significant progress in natural language processing.
\newcite{collobert:2011b} proposed a general framework which derives task-oriented
features by learning from raw text data using convolutional neural networks (CNN).
The idea of `learning from scratch' is fundamentally different from the conventional
methods which require careful and tedious feature engineering.
\newcite{collobert:2011b} evaluated the learning-based approach on several NLP tasks
including POS tagging, NER and semantic role labelling.
Without any human-designed features, they obtained close to or even better performance
than the state-of-the-art systems that involve complicated feature engineering.

A multitude of researches have been proposed to apply the deep learning methods and neural
models to relation classification. Most representative progress was made by \newcite{zeng14},
who proposed a CNN-based approach that can deliver quite competitive results without
any extra knowledge resource and NLP modules. Following the success of CNN,
there are some valuable models  such as multi-window CNN\cite{nguyenrelation},
CR-CNN\cite{dos2015classifying} and NS-depLCNN\cite{xu2015semantic} been proposed recently,
 which are all based on CNN structure. Though some models also based on other structures
like MV-RNN\cite{SocherEtAl2012:MVRNN}, FCM\cite{yufactor14} and SDP-LSTM\cite{xu2015classifying},
CNN occupies a leading position.

Despite the success obtained so far, most of the current CNN-based learning
approaches to relation learning and classification are weak in modeling temporal patterns.
Though SDP-LSTM algorithm utilize recurrent structure on dependency parsing,
no further analysis has been shown to compare RNN with CNN models.
Note that the semantic meaning of a relation is formed in the context of
the two target nominals, including the word sequence between them
and a window of preceding and following words.
Additionally, the relation is in fact `directional',
which means the order of the context words does matter.
Therefore, relation learning is essentially a task of temporal sequence learning,
and so should be modelled by a temporal model.
CNN models are static models, and are potentially weak especially
when learning long-distance relation patterns.
For example, the CNN model can learn only local patterns, and so is hard to deal with
patterns that is outside of the window of the convolutional filter. Dependency path can
alleviate this problem by removing noise compared with natural sequence input
(for example NS-depLCNN\cite{xu2015semantic}), but the computation cost and
adding error caused by dependency parser is inevitable. And the same limitation
still exists when the dependency path is long.

\begin{figure*}[!htb]
%\begin{figure*}[bp]
\centering
\includegraphics[height=7.4cm,width=11.6cm]{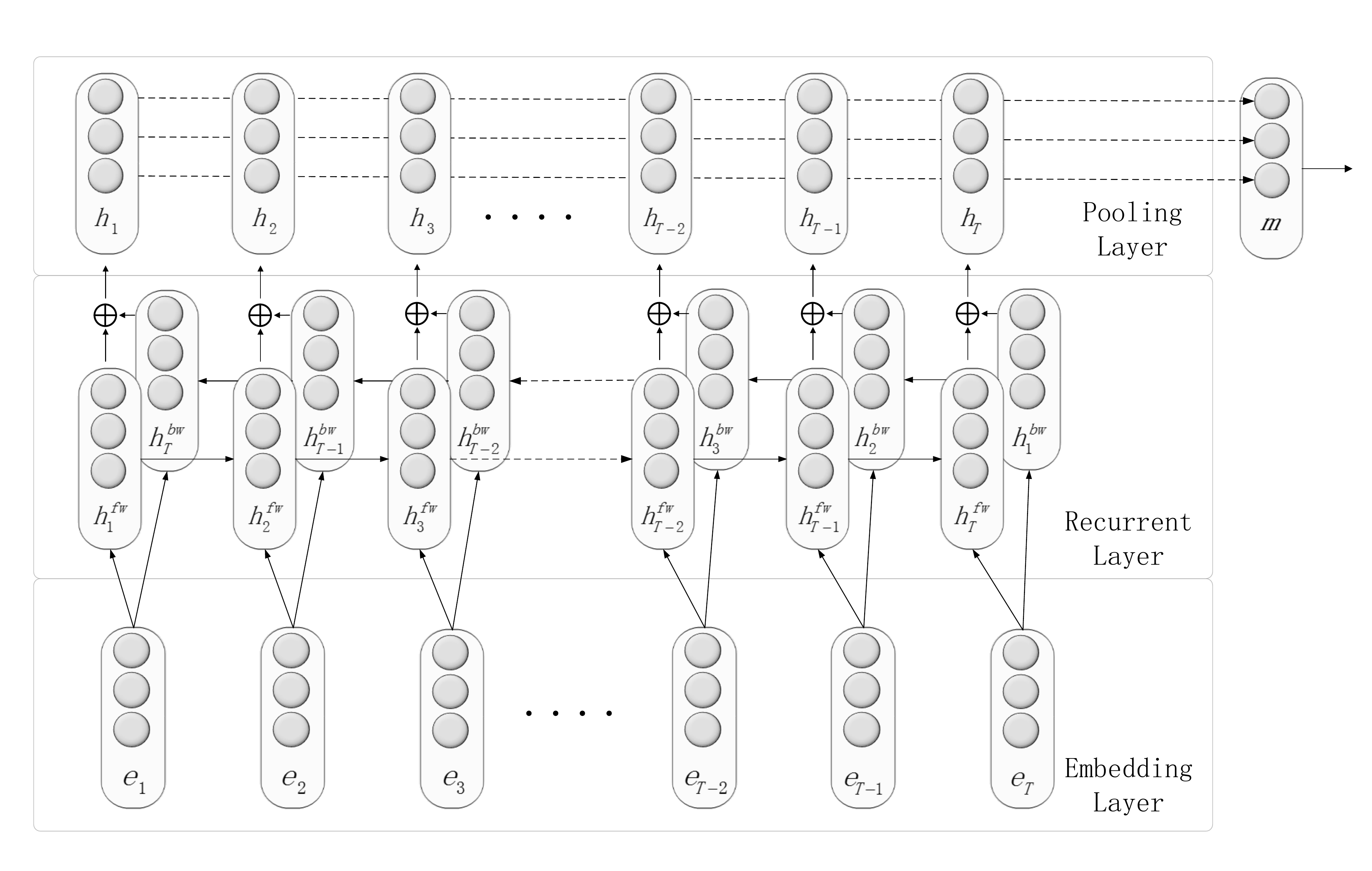}
\caption{The framework of the proposed model. }
\label{fig:framework}
\end{figure*}

In this paper, we propose a simple framework based on recurrent neural networks (RNN)
to tackle the problem of long-distance pattern learning. Compared to CNN models,
RNN is a temporal model and is particularly good at modeling sequential data~\cite{boden2001guide}.
The main framework is shown in Figure~\ref{fig:framework},
which will be described in details in Section \ref{sec:model}.

The main contributions of this paper are as follows:

\begin{itemize}

\item Proposed an RNN-based framework to model long-distance relation patterns.

\item Verified the advantages of recurrent structure not only on the most used SemEval-2010 task 8 dataset,
      but also on a new dataset refined from MIML-RE's annotated data\cite{2014emnlp-kbpactivelearning},
      and obtained distinct gain compared with CNN-based model.

\item Shows that position feature(PF) proposed by Zeng at al.\shortcite{zeng14} is less universal than
      position indicator(PI).

\item Analyzed empirically the capability of the RNN-based approach in modeling long-distance patterns.

\end{itemize}

\section{Related Work}

As mentioned, the conventional approaches to relation classification are based on pattern matching,
and can be categorized into feature-based methods~\cite{kambhatla2004combining,suchanek2006combining}
and kernel-based methods~\cite{bunescu2005shortest,mooney2005subsequence}.
The former category relies on human-designed patterns and so require
expert experience and time consuming, and the latter category suffers from data sparsity.
Additionally, these methods rely on extra NLP tools to derive linguistic features.

%The distant supervision methods can partially alleviate the difficulty in pattern design~\cite{mintz2009distant,riedel2010modeling,hoffmann2011knowledge,surdeanu2012multi},

%Move the part from introduction to here ?
To alleviate the difficulties in pattern design and also the lack of annotated data,
distant supervision has drawn a lot of attention since 2009~\cite{mintz2009distant,riedel2010modeling,hoffmann2011knowledge,surdeanu2012multi,2014emnlp-kbpactivelearning}.
This technique combines resources of text data and knowledge graph,
and uses the relations in the knowledge graph to discover patterns automatically from text data.

Our work follows the line of automatic feature learning by neural models,
which is largely fostered by~\newcite{collobert:2011b}.
A closely related work was proposed by~\newcite{zeng14},
which employed CNN to learn patterns of relations
from raw text data and so is a pure feature learning approach.
A potential problem of CNN is that this model can
learn only local patterns, and so is not suitable for learning
long-distance patterns in relation learning.
Particularly, simply increasing the window size of the convolutional
filters does not work: that will lose the strength of CNNs in modeling
local or short-distance patterns. To tackle this problem, \newcite{nguyenrelation}
proposed a CNN model with multiple window sizes for filters, which allows learning
patterns of different lengths. Although this method is promising,
it involves much more computation, and tuning the window sizes is not trivial.
The RNN-based approach could solve the difficulty of CNN models
in learning long-distance and variable-distance patterns in an elegant way.

In order to tackle with long-distance dependency patterns,
some works are proposed based on dependency trees which can eliminate irrelevant words in the sentence.
One early work is MV-RNN model proposed by~\newcite{SocherEtAl2012:MVRNN}.
The difference is that we based on different RNNs: the MV-RNN model is based on recursive
NN while our work is based on recurrent NN, a temporal model.
Recently, Kun Xu at al.~\shortcite{xu2015semantic} exploits the dependency path to learn
the assignments of subjects and objects using a straightforward negative sampling method,
which adopts the shortest dependency path from the object to the subject as a negative sample.
More recently, Yan Xu at al. ~\shortcite{xu2015classifying} proposed a model based on
LSTM recurrent neural network, which is most similar with our model.
However, these works rely on syntactic parsing, which makes the process more complicated.
When sentence becomes longer and the syntax becomes more complex,
more error from dependency tree will appear thus influence the final performance.

%Unlike previous works, our model uses only word vectors
%and so is more efficient especially in predicting process.

%Thus, a natural idea comes, the RNN-based approach could solve the difficulty of CNN models
%in learning long-distance and variable-distance patterns in an elegant way. While there is no successful
%achievement so far, one reason is that key words and phrases within three words can cover most of the short patterns
%related to relation classification. But when those key phases are composed into complicated grammar or long sentences,
%it is difficult to tell which key phases are irrelevant. And the position feature~\cite{zeng14} can only partly solve
%the problem. This leads to another reason, the dataset itself is quite limited.

In addition, our work is related to the FCM framework~\cite{yufactor14}.
In principle, FCM decomposes sentences into substructures and factorizes
semantic meaning into contributions from multiple annotations (e.g., POS, NER, dependency parse).
It can be regarded as a general form of the MV-RNN and CNN models where the recursive hierarchy
or max-pooling are replaced by a general composition function. Nevertheless,
FCM is still a static model and shares the same disadvantage of CNN in modeling temporal data.
~\newcite{dos2015classifying} also use the convolutional network.
And they propose a ranking-based cost function and elaborately diminish the impact of the Other class.

The advantage of the RNN model in learning sequential data is well-known and has been utilized in language
modeling~\cite{mikolov2010recurrent} and sequential labeling~\cite{schuster1997bidirectional}.
Compared to these studies, a significant difference of our model is that there are no
predicting targets at each time step, and the supervision (relation label) is only available at the end of a sequence.
This is similar to the semantic embedding model proposed by~\newcite{palangi2015deep}, though we have made
several important modifications, as will be presented in the next section.

\section{Model}
\label{sec:model}

As has been shown in Figure~\ref{fig:framework}, the model proposed in
this paper contains three components:
(1) a word embedding layer that maps each word in a sentence
into a low dimension word vector;
(2) a bidirectional recurrent layer that models the
word sequence and produces word-level features (representations);
(3) a max pooling layer that merges word-level features from each time step
(each word) into a sentence-level feature vector, by selecting the maximum value
among all the word-level features for each dimension.
The sentence-level feature vector is finally used for relation classification.
These components will be presented in detail in this section.

\subsection{Word embedding}

The word embedding layer is the first component of the proposed model,
which projects discrete word symbols to low-dimensional dense word vectors,
so that the words can be modeled and processed by the following layers.
Let $x_t \in \{0,1\}^{|V|}$ denote the one-hot representation of the $t$-th word $v_t$,
where $|V|$ denotes the size of the vocabulary $V$.
The embedding layer transfers $x_t$ to word vectors $e_t \in R^{D}$ as follows:

\begin{equation}
\label{eq:embed}
e_t = W_{em} x_t
\end{equation}

\noindent where $W_{em} \in R^{|D| \times |V|}$ is the projection matrix.
Since $x_t$ is one-hot, $W_{em}$ in fact stores representations of all the words in $V$.
Word embedding has been widely studied in the context of semantic learning.
In this work, we first train word vectors using the word2vec
tool\footnote{\url{http://code.google.com/p/word2vec/}} with a large amount of data
that are in general domains, and then use these vectors to initialize (pre-train)
the word embedding layer of our model. By this way, knowledge of general domains can be used.
It has been shown that this pre-training improves model training, e.g., ~\cite{zeng14,yufactor14}.

\subsection{Bi-directional network}

The second component of our model is the recurrent layer, the key part for modeling
sequential data and long-distance patterns. We start from a simple one-directional
forward RNN. Given a sentence $X=(x_1, x_2, ..., x_T)$, the words are projected into a
sequence of word vectors, denoted by $(e_1, e_2, ..., e_T)$ where $T$ is the number of words.
These word vectors are put to the recurrent layer step by step. For each step $t$, the
network accepts the word vector $e_t$ and the output at the previous step $h_{t-1}^{fw}$
as the input, and produces the current output $h_{t}^{fw}$ by a linear transform
followed by a non-linear activation function% not smooth
, given by:

\begin{equation}
\label{eq:forward}
h_{t}^{fw} = \tanh (W_{fw} e_t + U_{fw} h_{t-1}^{fw} + b_{fw})
\end{equation}

\noindent where $h_{t}^{fw} \in R^{M}$ is the output of the RNN at the $t$-th step,
which can be regarded as local segment-level features produced by the word segment $(x_1,...,x_t)$.
Note that $M$ is the dimension of the feature vector, and $W_{fw} \in R^{M \times D}$,
$U_{fw} \in R^{M \times M} $, $b_{fw} \in R^{M \times 1 } $ are the model parameters.
We have used the hyperbolic function $\tanh(\cdot)$ as the non-linear transform,
which can help back propagate the error more easily due to its symmetry~\cite{glorot2010understanding}.

A potential problem of the one-directional forward RNN is that the information of future words
are not fully utilized when predicting the semantic meaning in the middle of a sentence.
A possible solution is to use a bi-directional architecture that
makes predictions based on both the past and future words, as has been seen in Figure.~\ref{fig:framework}.
This architecture has been demonstrated to work well in sequential labeling, e.g.,~\cite{schuster1997bidirectional}.
With the bi-directional RNN architecture, the prediction at step $t$ is obtained by simply
adding the output of the forward RNN and the backward RNN, formulated as follows:

\begin{equation}
\label{eq:add}
h_t = h_{t}^{fw} + h_{t}^{bw}
\end{equation}

\noindent where $h_{t}^{bw} \in R^{M \time 1}$ is the output of the backward RNN,
which possesses the same dimension as $h_{t}^{fw}$ defined by:

\begin{equation}
\label{eq:backward}
h_{t}^{bw} = \tanh (W_{bw} e_t  +  U_{bw} h_{t+1}^{bw} + b_{bw})
\end{equation}

\noindent where $W_{bw} \in R^{M \times D } $, $U_{bw} \in R^{M \times M}$,
$b_{bw} \in R^{M \times 1 }$  are the parameters of the backward RNN.
Note that the forward and backward RNNs are trained  simultaneously,
and so the addition is possible even without any parameter sharing between the two RNN structures.

\subsection{Max-pooling}

Sentence-level relation classification requires a single sentence-level feature vector
to represent the entire sentence. In the CNN-based models, a pooling approach is often
used~\cite{zeng14}. With the RNN structure, since the semantic meaning of a sentence is
learned word by word, the segment-level feature vector produced at the end of the sentence actually
represents the entire sentence. This accumulation approach has been used in ~\cite{palangi2015deep}
for sentence-level semantic embedding.

In practice, we found that the accumulation approach is not very suitable for relation learning because
there are many long-distance patterns in the training data. Accumulation by recurrent connections
tends to forget long-term information quickly, and the supervision at the end of the
sentence is hard to be propagated to early steps in model training, due to the annoying
problem of gradient vanishing~\cite{bengio1994learning}. %{\color{red}Particularly, with the bi-directional
%architecture, the prediction at step $t$ depends on both the past and future words
%breaks the temporal accumulation property and so the accumulation approach does not work.}

We therefore resort to the max-pooling approach as in CNN models. The argument is that the
segment-level features, although not very strong in representing the entire sentence, can represent local
patterns well. The semantic meaning of a sentence can be achieved by merging representations
of the local patterns. The max-pooling is formulated as follows:

\begin{equation}
\label{eq:max}
m_i = \max_{t} \{ (h_t)_i \} , \ \ \ \forall i=1,...,M
\end{equation}

\noindent where $m$ is the sentence feature vector and $i$ indexes feature dimensions.

Note that we have chosen max-pooling rather than mean-pooling. The hypothesis is that only
several key words (trigger) and the associated patterns are important for relation classification,
and so max-pooling is more appropriate to promote the most informative patterns.

\subsection{Model training}

Training the model in Figure~\ref{fig:framework} involves optimizing the parameters
$\theta=\{W_{in}, W_{fw}, U_{fw}, b_{fw}, W_{bw},$ $ U_{bw}, b_{bw}\}$. The training objective is
that, for a given sentence, the output feature vector $h$ achieves the best performance
on the task of relation classification.
%This optimization largely depends on the classifier, %why mention this?
Here we use a simple logistic regression model as the classfier.
Formally, this model predicts the posterior probability that an input sentence $X$
involves a relationship $r$ as follows:

\begin{equation}
P(r|X, \theta, W_o, b_o) = \sigma (W_o h(X) + b_o)
\end{equation}
\noindent where $\sigma(x) = \frac{e^{x_i}}{\sum_{j} e^{x_j}}$ is the softmax function,
and $\theta$ encodes the parameters of the RNN model.

Based on the logistic regression classifier, a natural objective function is the cross
entropy between the predictions and the labels, given by:

\begin{equation}
\label{eq:cost}
\mathcal{L} (\theta, W_o, b_o) =  \sum_{n \in N} - \log p(r^{(n)}| X^{(n)},\theta, W_o, b_o)
\end{equation}

\noindent where $n$ is the index of sentences in the training data,
and $X^{(n)}$ and $r^{(n)}$ denote the $n$-th sentence and its relation label, respectively.

To train such a model, we follow the training method proposed by~\newcite{collobert:2011b},
and utilizes the stochastic gradient descent (SGD) algorithm. Specifically,
the back propagation through time (BPTT)~\cite{werbos1990backpropagation}
is employed to compute the gradients layer by layer, and the fan-in technique
proposed by~\newcite{plaut1987learning} is used to initialize the parameters.
It was found that this initialization
can locate the model parameters around the linear region of the activation function,
which helps propagating the gradients back to early steps easier. Moreover, it
also balances the learning speed for parameters in different layers~\cite{lecun2012efficient}.

As has been discussed, pre-training the word embedding layer with word vectors
trained from extra large amount corpus improves the performance. This approach has been
employed in our experiments.

\subsection{Position indicators}

In relation learning, it is essential to let the algorithm know the target nominals.
In the CNN-based approach, \newcite{zeng14} appended a position feature vector to
each word vector, i.e., the distance from the word to the two nominals. This has
been found highly important to gain high classification accuracy and some works have followed
this technique\cite{nguyenrelation,dos2015classifying}.
For RNN, since the model learns the entire word sequence, the \emph{relative} positional information
for each word can be obtained automatically in the forward or backward recursive propagation.
It is therefore sufficient to annotate the target nominals in the word sequence, without
necessity to change the input vectors.

We choose a simple method that uses four position indicators(PI) to specify the starting and ending
of the nominals. The following is an example: ``$<$e1$>$ {\bf people} $<$/e1$>$ have
been moving back into $<$e2$>$ {\bf downtown} $<$/e2$>$''. Note that
{\bf people} and {\bf downtown} are the two nominals with the relation `Entity-Destination(e1,e2)',
and $<$e1$>$, $<$/e1$>$, $<$e2$>$, $<$/e2$>$ are
the four position indicators which are regarded as single words in the training and testing process.
The position-embedded sentences are then
used as the input to train the RNN model. Compared to the
position feature approach in the CNN model, the position indictor method is
more straightforward. And in section \ref{sec:result}, experiment results show that under most circumstances,
PI can be more helpful than PF.

\section{Experiments}

\subsection{Database}

\begin{table*}[!htb]
\begin{center}
\begin{tabular}{|l|c|c|}
\hline
Number of training data     &  15917 \\
Number of development data      &  1724  \\
Number of test data         &  3405  \\
\hline
Number of relation types          &  37    \\
\hline
per:alternate\_names               &  org:alternate\_names  \\
per:origin                        &  org:subsidiaries \\
per:spouse                        &  org:top\_members/employees \\
per:title                         &  org:founded \\
per:employee\_of                   &  org:founded\_by \\
per:countries\_of\_residence        &  org:country\_of\_headquarters \\
per:stateorprovinces\_of\_residence &  org:stateorprovince\_of\_headquarters \\
per:cities\_of\_residence           &  org:city\_of\_headquarters \\
per:country\_of\_birth              &  org:members   \\
no\_relation                       &                    \\
\hline
\end{tabular}
\end{center}
\caption{\label{tab:KBP37} Statistics of KBP37 .}
\end{table*}

We use two different datasets.
The first one is the dataset provided by SemEval-2010 Task 8.
There are $9$ \emph{directional} relations and an additional `other' relation,
resulting in $19$ relation classes in total. Given a sentence and two target nominals,
a prediction is counted as correct only when both the relation and its direction are correct.
The performance is evaluated in terms of the F1 score defined by SemEval-2010 Task 8~\cite{hendrickx2009semeval}.
Both the data and the evaluation tool are publicly available.
\footnote{\url{http://docs.google.com/View?docid=dfvxd49s_36c28v9pmw}}

The second dataset is a revision of MIML-RE annotation dataset,
provided by Gabor Angeli et al.~\shortcite{2014emnlp-kbpactivelearning}.
They use both the 2010 and 2013 KBP official document collections, as well as a July 2013 dump
of Wikipedia as the text corpus for annotation. There are 33811 sentences been annotated.
To make the dataset more suitable for our task, we made several refinement:

\begin{enumerate}

\item First, we add direction to the relation names, such that `per:employee\_of' is splited into
two relations `per:employee\_of(e1,e2)' and `per:employee\_of(e2,e1)' except for `no\_relation'.
According to description of KBP task,\footnote{\url{http://surdeanu.info/kbp2013/TAC_2013_KBP_Slot_Descriptions_1.0.pdf}}
we replace `org:parents' with `org:subsidiaries' and replace `org:member\_of' with `org:member'
(by their reverse directions). This leads to 76 relations in the dataset.

\item Then, we statistic the frequency of each relation with two directions separately.
      And relations with low frequency are discarded so that both directions of each relation
      occur more than 100 times in the dataset. To better balance the dataset,
      80\% `no\_relation' sentences are also randomly discarded.

\item After that, dataset are randomly shuffled and then sentences under each relation are
      all split into three groups, 70\% for training, 10\% for development, 20\% for test.
      Finally, we remove those sentences in the development and test set whose entity pairs
      and relation are appeared in a training sentence simultaneously.

\end{enumerate}

In the rest of this paper, we will call the second dataset KBP37 for the sake of simplicity.
Statistics and relation types are shown in Table \ref{tab:KBP37}. KBP37 contains $18$ \emph{directional}
relations and an additional `no\_relation' relation, resulting in $37$ relation classes.
Notice that KBP37 is different from SemEval-2010 Task 8 in several aspects:
\begin{itemize}

\item Pairs of nouns in KBP37 are always entity names which are more sparse than SemEval-2010 task 8.
      And there are more target nouns that own several words instead of one, which is barely unseen in previous dataset.
\item Average length of sentences in KBP37 is much longer than Semeval-2010 task 8, which will be discussed in details
      in section \ref{sec:proportion_of_long_context}
\item It is not guaranteed that there exists only one relation per data, though in test set only one
      relation is offered as the answer.

\end{itemize}
The last aspect may lead to some inconsistency from our task. But since multi-relation data rarely exists,
it can be omitted.

\subsection{Experimental setup}

In order to compare with the work by~\newcite{SocherEtAl2012:MVRNN} and~\newcite{zeng14},
we use the same word vectors proposed by~\newcite{turian2010word} (50-dimensional) to initialize
the embedding layer in the main experiments. Additionally, to compare with other recent models,
additional experiments are also conducted with the word vectors
pre-trained by~\newcite{mikolov2013distributed} which are 300-dimensional.

Because there is no official development dataset in Semeval-2010 Task 8 dataset,
we tune the hyper-parameters by 8-fold cross validation. % training data(7001-8000).
Once the hyper-parameters are optimized, all the 8000 training data are used to train the
model with the best configuration. With Turian's 50-dimensional word vectors,
the best dimension of the feature vector $m$ is $M=800$,  and with Mikolov's 300-dimensional word vectors,
the best feature dimension is $M=$. The learning rate is set to be $0.01$ in both two conditions.
For fast convergence, we set learning rate to $0.1$ in the first five iterations. And iteration time is 20.

And since we split a development set for KBP37, it is more convenient to tune the hyper-parameters.
In the test process, development data is also used to choose the best model among different iterations.
For KBP37, we only use Turian's 50-dimensional word vectors. The best dimension of feature vector $m$ is
$M=700$. And max iteration time is 20.

\subsection{Results}
\label{sec:result}

\begin{table}[!htb]
\begin{center}
\begin{tabular}{|l|c|c|}
\hline
Model                                 & F1 \\
\hline
RNN                                   & 31.9\\
+ max-pooling                         & 67.5\\
+ position indicators                 & 76.9\\
+ bidirection                         &  79.6\\
\hline
\end{tabular}
\end{center}
\caption{\label{tab:contribution} F1 results with the proposed RNN model on SemEval-2010 Task 8 dataset,
          $+$ shows the performance after adding each modification. }
\end{table}

%\begin{table}[!htb]
%\begin{center}
%\begin{tabular}{|l|l|c|c|}
%\hline
%Model              & Features               & F1 \\
%\hline
%MV-RNN             & syntactic parse             & 79.1  \\
%\cite{SocherEtAl2012:MVRNN} && \\
%CNN                & PF    & 78.9 \\
%\cite{zeng14}& & \\
%\hline
%RNN (proposed)     & PI                     &\bf79.6 \\
%\hline
%\end{tabular}
%end{center}
%\caption{\label{tab:bestresult} Comparison of F1 scores with different neural models.
%The 50-dimensional word vectors provided by~\newcite{turian2010word}
%are used for pre-training. PF stands for position features and PI stands for position indicators.}
%\end{table}

\begin{table}[!htb]
\begin{center}
\begin{tabular}{|l|l|l|}
\hline
Model              & Semeval-2010                    & KBP37 \\
                   & task8                           &     \\
\hline
MV-RNN             & 79.1 & - \\
(Socher, 2012)     & & \\
CNN+PF             & 78.9 & - \\
(Zeng, 2014)       & &  \\

\hline
CNN+PF             & 78.3                            & 51.3\\
(Our)              & ($300\rightarrow300$)           & ($500\rightarrow500$) \\
CNN+PI             & 77.4                            & 55.1 \\
(Our)              & ($400\rightarrow400$)           & ($500\rightarrow500$) \\
\hline
RNN+PF             & 78.8($400$)                     & 54.3($400$)   \\
RNN+PI             & \bf79.6($800$)                  &\bf58.8($700$) \\
\hline
\end{tabular}
\end{center}
\caption{\label{tab:twodataset} Comparing F1 scores with different neural models,
different position usage and different datasets.
The 50-dimensional word vectors provided by~\newcite{turian2010word}
are used for pre-training. PF stands for position features and PI stands
for position indicators. Number in the parentheses shows the  best dimension of hidden layer.}
\end{table}

Table~\ref{tab:contribution} presents the F1 results of proposed RNN model,
with the contribution offered by each modification. It can be seen that the basic RNN,
which is signal directional and with the output of the last step as the sentence-level features,
performs very poor. This can be attributed to the lack of the position information
of target nominals and the difficulty in RNN training. The max-pooling offers
the most significant performance improvement, indicating that local patterns learned
from neighbouring words are highly important for relation classification.
The position indicators also produce highly significant improvement, which is not
surprising as the model would be puzzled which pattern to learn without the positional information.
The contribution of positional information has been demonstrated by~\newcite{zeng14},
where the positional features lead to nearly $10$ percentiles of F1 improvement,
which is similar as the gain obtained in our model.

The second experiment compares three representative neural models with different datasets and
different position information added. The results are presented in Table~\ref{tab:twodataset}.
The 50-dimensional word vectors are employed in this table.
In our experiments, though rather close, we didn't reproduce 78.9 of F1 value reported by ~\newcite{zeng14},
the third and fourth column shows our results with CNN-based model.
%To implement PF, we randomly initialize 400 vectors with dimension 5 and range 0.02 to represent
%distance between current input word and first words of noun pair. And
%the PF vectors are also updated during training process.
Compared among different rows, we can come to the conclusion that the RNN model outperforms
both the MV-RNN model proposed by~\cite{SocherEtAl2012:MVRNN} and the CNN model proposed by~\cite{zeng14}.
Compared among different columns, the results show that RNN model obtains more
improvement comparing to CNN on KBP37 dataset, which also indicates the difference between two datasets.
More discussion will be held in section \ref{sec:proportion_of_long_context}.

From Table~\ref{tab:twodataset}, we can draw another conclusion that, with recurrent structure,
PI is more effective than PF. And PI contributes even more when experiments are implemented on KBP37 dataset.
The former phenomenon should caused by the ambiguous accumulation in the recurrent process which may disrupt
the PF information. The latter probably attributes to a shortcoming of PF that it tends to be less accurate
when the nouns become phrases instead of single words.
More discussion will be held in Section \ref{sec:impact_of_long_context}

An interesting observation is that the RNN model
performs better than the MV-RNN model which uses syntactic parse as extra resources.
This indicates that relation patterns can be effectively
learned by RNNs from raw text, without any explicit linguistic knowledge.

%Finally, we compare the RNN model with several representative models that achieve
%state-of-the-art results in relation classification (see Table \ref{tab:result}).
%The first model is based on SVMs and was proposed by~\newcite{hendrickx2009semeval}.
%This model can represent the state-of-the-art pattern-based system.
%All the other models are based on neural networks, which are MV-RNN~\cite{SocherEtAl2012:MVRNN},
%CNN~\cite{zeng14}, CNN with multiple window sizes~\cite{nguyenrelation}, FCM~\cite{yufactor14},
%ranking based classification model~\cite{dos2015classifying}, LSTM-based model with dependency path~\cite{xu2015classifying}
%and CNN-based model with negative sampling over dependency path~\cite{xu2015semantic}.
%Note that different authors use different features and extra knowledge,
%which makes it difficult to compare the results directly.
%Among the learning-from-scratch models,
%i.e., no extra NLP processing involved, the best performance ($84.1$)
%is achieved by the ~\newcite{dos2015classifying}. Compared to CNN model\cite{zeng14},
%our RNN model achieves a similar result ($82.5$) without wordnet information,
%and the network structure is simpler.
In the next section, we will show that the RNN model possesses more
potential in real application with complex long-distance relations.

\section{Discussion}

\subsection{Impact of long context}
\label{sec:impact_of_long_context}

We have argued that a particular advantage of the RNN model compared to the CNN model is that
it can deal with long-distance patterns more effectively. To verify this argument,
we split the test datasets into $5$ subsets according to length of the context.
Here the context is defined as words between the two nominals plus $3$ words prior
to the first nominal and $3$ words after the second nominal, if they exist. The position
indicator does not count. Clearly, long contexts lead to long-distance patterns.
In order to compare performance of the RNN and CNN models, we produce the CNN-based method with PI.
This modification ensures that the two models learn the same input sequence with
the same representation.

\begin{figure}
\begin{minipage}[t]{0.5\linewidth}
\centering
\includegraphics[width=1.5in]{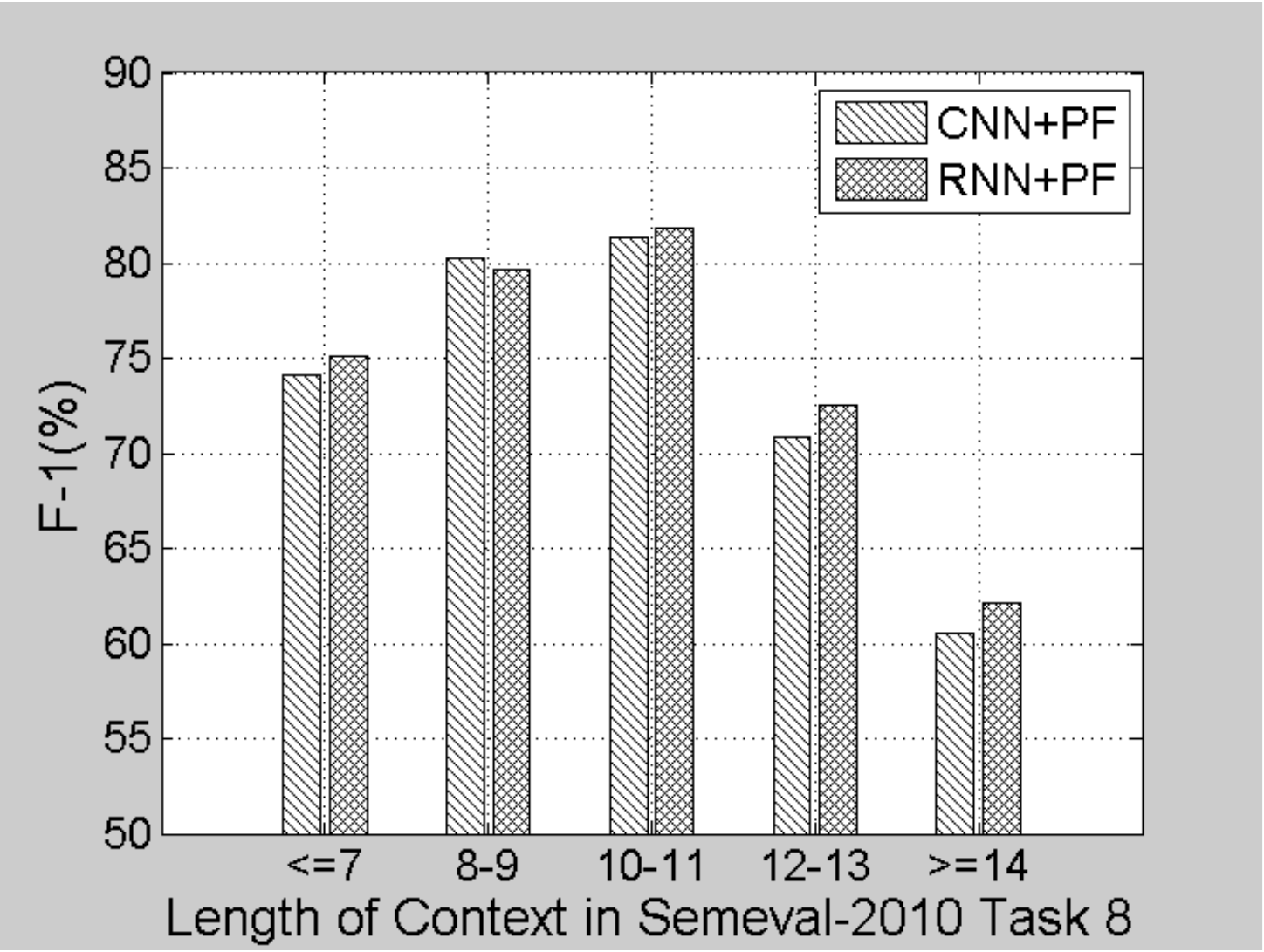}
\end{minipage}%
\begin{minipage}[t]{0.5\linewidth}
\centering
\includegraphics[width=1.5in]{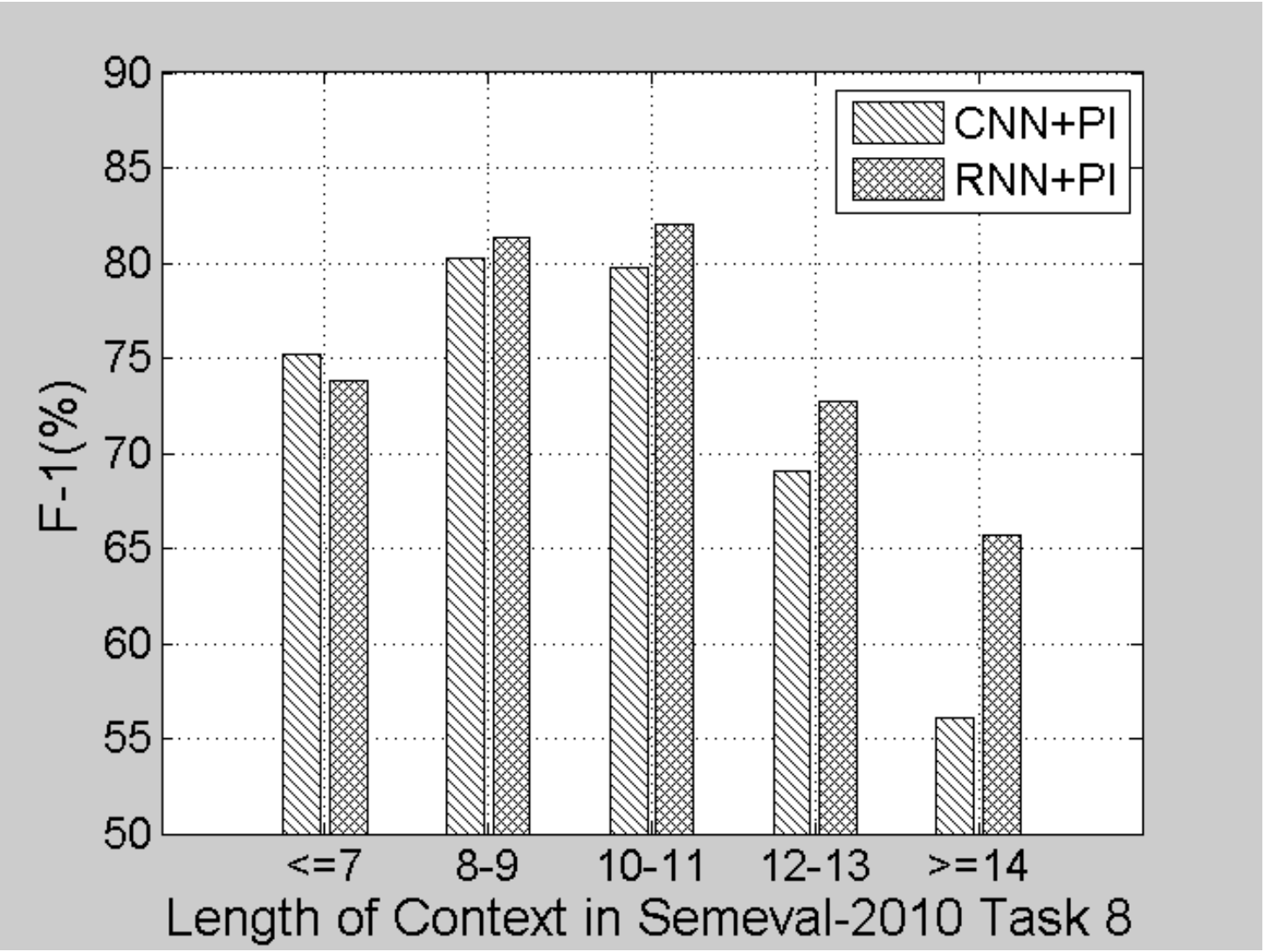}
\end{minipage}

\begin{minipage}[t]{0.5\linewidth}
\centering
\includegraphics[width=1.5in]{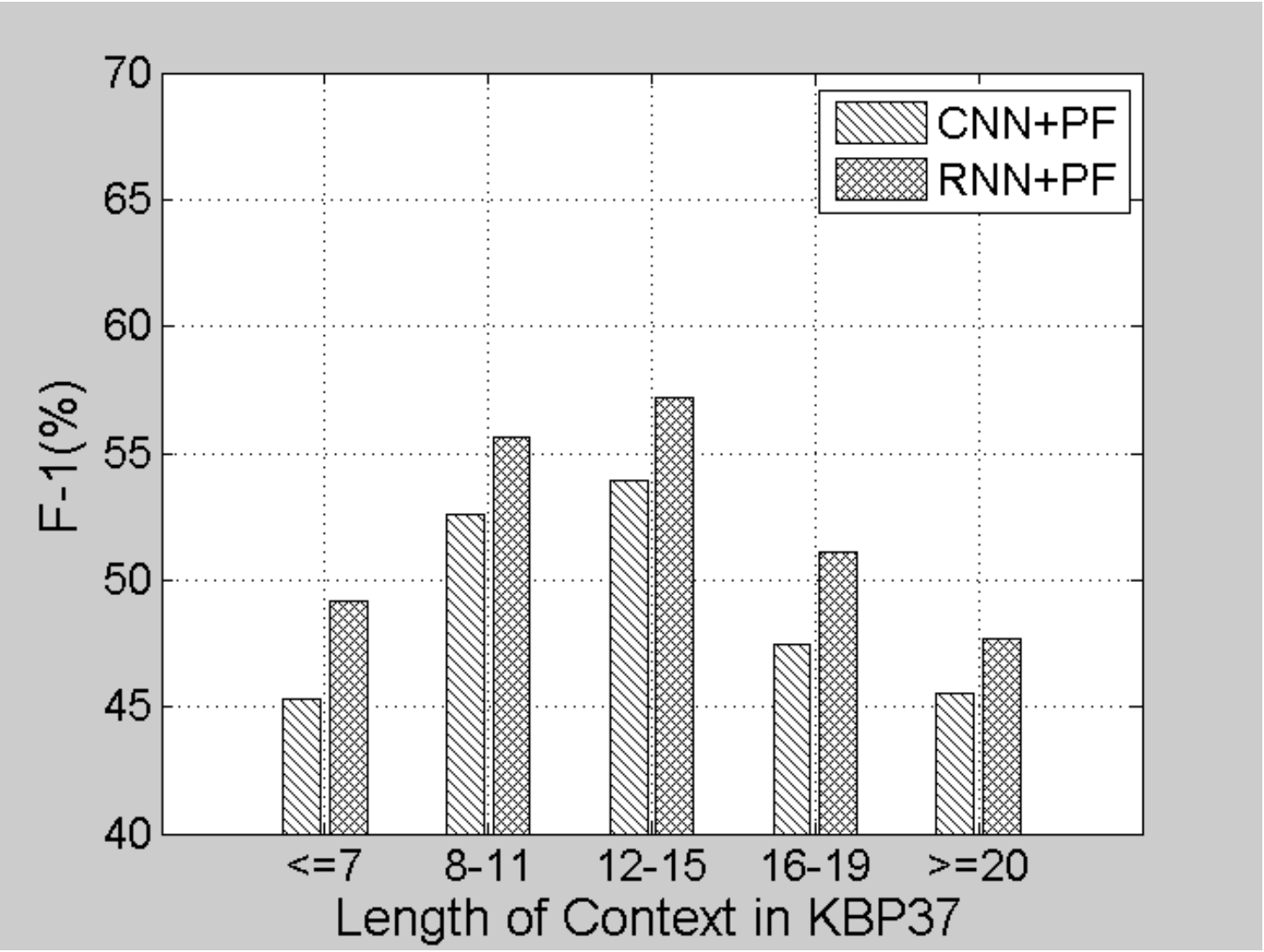}
\end{minipage}%
\begin{minipage}[t]{0.5\linewidth}
\centering
\includegraphics[width=1.5in]{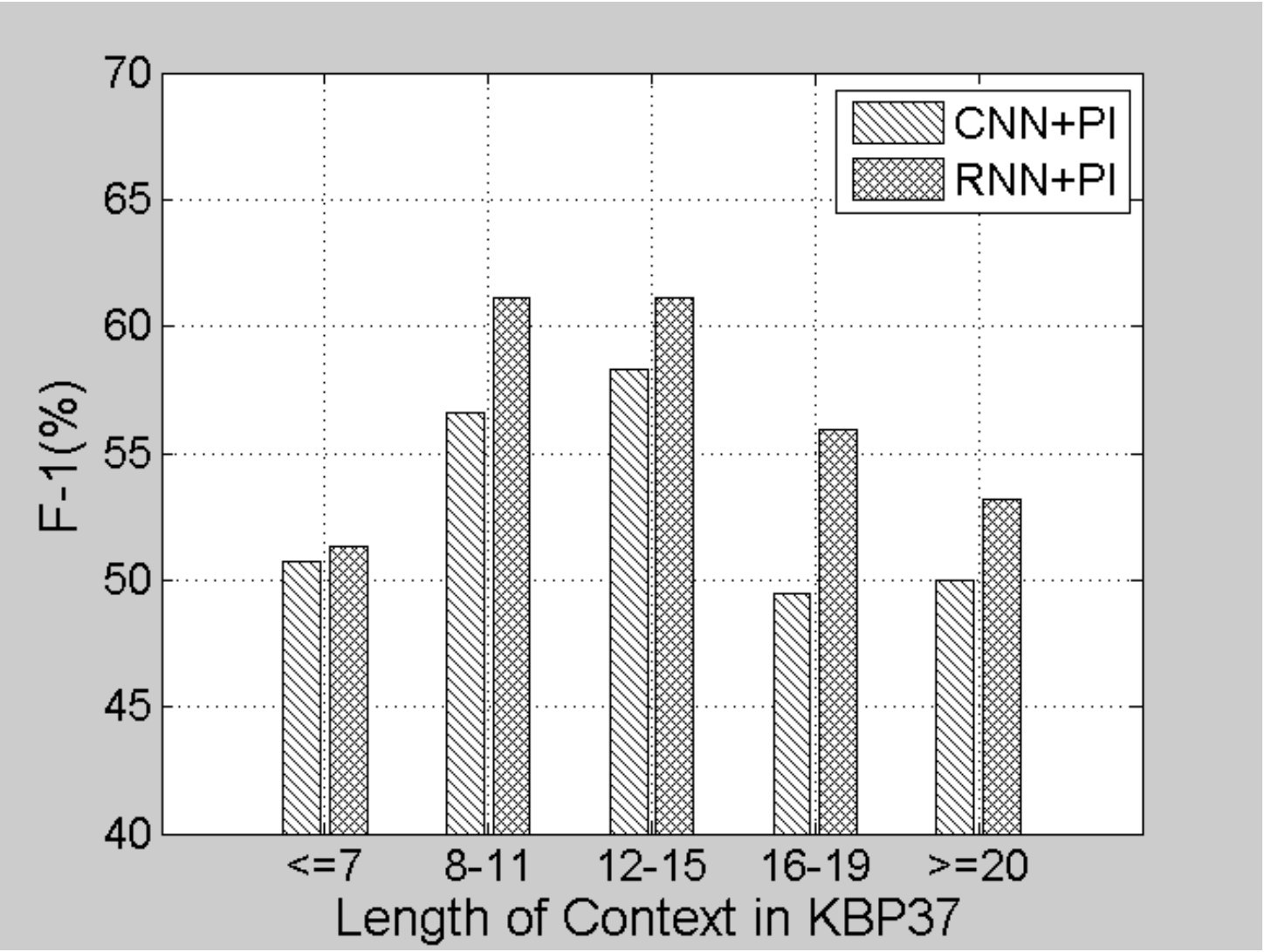}
\end{minipage}
\caption{F1 scores with different length of contexts in Semeval-2010 Task 8 and KBP37. }
\label{fig:cluster}
\end{figure}

The F1 results on the $5$ subsets are reported in
Figure~\ref{fig:cluster}. It can be seen that if the context length is small,
the CNN and RNN models perform similar with PI, whereas if the context length is large,
the RNN model is clearly superior. This confirms that RNN is more suitable to learn long-distance patterns.
The results also shows that when PF is used, there is no such clear trend,
which means PF is more suitable for CNN models and PI is more suitable for RNN models.
But since PI performs better than PF with CNN in KBP37(51.3 to 55.1 from Table \ref{tab:twodataset}),
PI could be more robust.

Note that with both two models, the best F1 results are obtained
with a moderate length of contexts. This is understandable as too small context
involves limited semantic information, while too large context leads to difficulties in pattern learning.
%\begin{figure}[!htb]
%\centering
%\epsfig{figure=length.eps,height=4.3cm,width=7cm}
%\caption{F1 scores with different length of contexts. }
%\label{fig:cluster}
%\end{figure}

%\begin{figure}[!htb]
%\centering
%\epsfig{figure=lengthsemeval_pf.eps,height=4.3cm,width=6.5cm}
%\epsfig{figure=lengthsemeval_pi.eps,height=4.3cm,width=6.5cm}
%\epsfig{figure=lengthkbp_pf.eps,height=4.3cm,width=6.5cm}
%\epsfig{figure=lengthkbp_pi.eps,height=4.3cm,width=6.5cm}
%\caption{F1 scores with different length of contexts in Semeval-2010 Task 8 and KBP37. }
%\label{fig:cluster}
%\end{figure}

\subsection{Proportion of long context }
\label{sec:proportion_of_long_context}

\begin{table*}[htb]
\begin{center}
\begin{tabular}{|l|c|c|c|c|}
\hline
\multirow{2}{*}{Dataset} & \multicolumn{3}{c|}{Context Length}            & Proportion of\\
\cline{2-4}                    &   $\le$10  & 11 - 15 & $\geq$ 16         & Long Context ($\geq 11$) \\
\hline
SemEval-2010 task-8~\cite{hendrickx2009semeval}     & 6658   & 3725  & 334  & 0.379  \\
NYT+Freebase~\cite{riedel2013relation}              & 22057  & 19369 & 3889 & 0.513  \\
KBP+Wikipedia~\cite{2014emnlp-kbpactivelearning}              & 6618   & 11647 & 15546 & 0.804 \\
\hline
\end{tabular}
\end{center}
\caption{\label{tab:contextlength} The distribution of context lengths with three datasets.}
\end{table*}

Figure~\ref{fig:cluster} shows that the RNN model significantly outperforms the CNN model,
which is a little different from the results presented in Table~\ref{tab:twodataset}, where the
discrepancy between the two models in SemEval-2010 dataset is not so remarkable ($77.4$ vs. $79.6$).
This can be attributed to the small proportion of long contexts in test data.

To reflect the limitation of SemEval-2010 dataset, the distribution of the context lengths is calculated
on the test dataset. For comparison, the New York Time corpus with the entities and relations
selected from a subset of Freebase recommended by~\newcite{riedel2013relation} is also presented.

%Note that for Limin Yao's dataset, we pick up all the `POSITIVE' sentences from both the training and test data,
%for which every sentence involves at least one pair of `related' entities in Freebase.

The statistics are shown in Table~\ref{tab:contextlength}. It can be observed that
long contexts exist in all the three datasets. Particularly, the proportion of
long contexts in the SemEval-2010 task 8 dataset is rather small compared to the other two datasets.
This suggests that the strengths of the different models were not fully demonstrated by only implementing
experiments on SemEval-2010 task 8 dataset. Since most recent works on relation classification only implemented
on this single dataset, a comparison among different models on KBP37 dataset is needed.

\subsection{Semantic accumulation}

Another interesting analysis is to show how the `semantic meaning' of a sentence
is formed. First notice that with both the CNN and the RNN models,
the sentence-level features are produced from local features (word-level for CNN and
segment-level for RNN) by dimension-wise max-pooling.

%\begin{figure*}[!htb]
%\centering
%\includegraphics[height=3.5cm,width=11cm]{sample_analysis2.pdf}
%\caption{Semantic distribution on words in the sentence ``A $<$e1$>$ witch $<$/e1$>$
%is able to change events by using $<$e2$>$ magic $<$/e2$>$ ."  }
%\label{fig:sentence1}
%\end{figure*}
%\begin{figure*}[!htb]
%\centering
%\includegraphics[height=3.7cm,width=16cm]{sample_analysis1.pdf}
%\caption{Semantic distribution on words in the sentence ``Skype, a free software, allows
%a $<$e1$>$ hookup $<$/e1$>$ of multiple computer $<$e2$>$ users $<$/e2$>$ to join in an
%online conference call without incurring any telephone costs."  }
%\label{fig:sentence2}
%\end{figure*}

\begin{figure}[!htb]
\centering
\includegraphics[height=2cm,width=5cm]{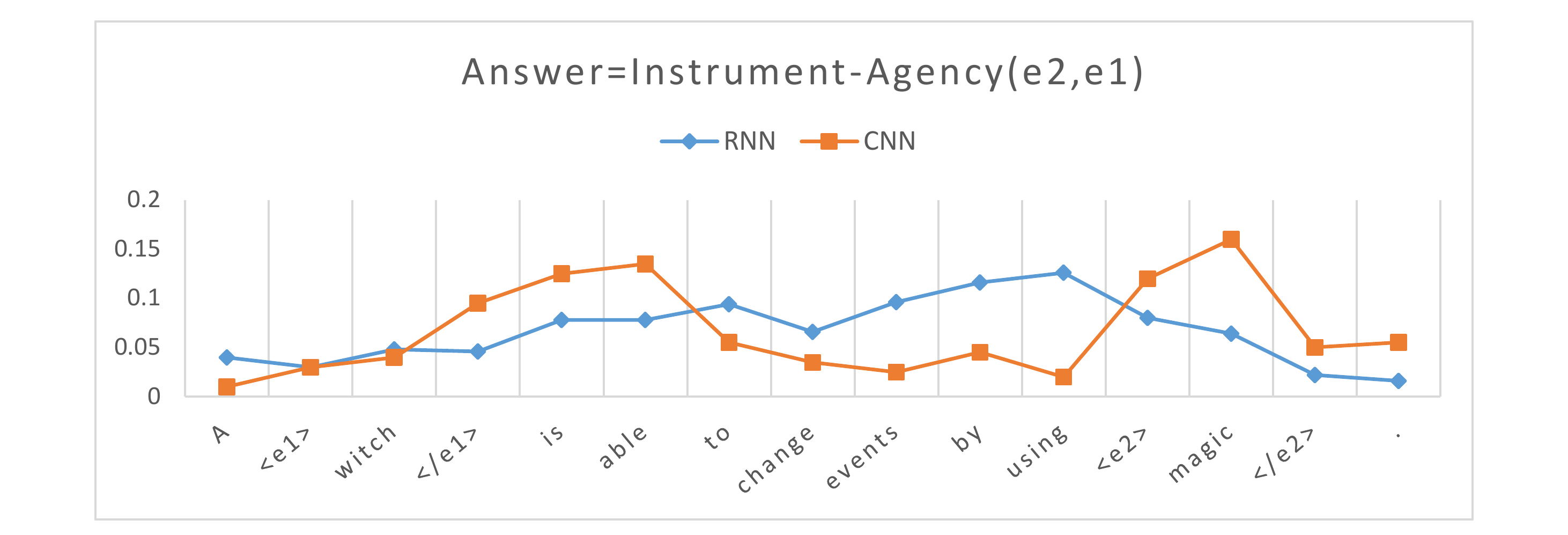}
\caption{Semantic distribution on words in the sentence ``A $<$e1$>$ witch $<$/e1$>$
is able to change events by using $<$e2$>$ magic $<$/e2$>$ ."  }
\label{fig:sentence1}
\end{figure}
\begin{figure}[!htb]
\centering
\includegraphics[height=2cm,width=8cm]{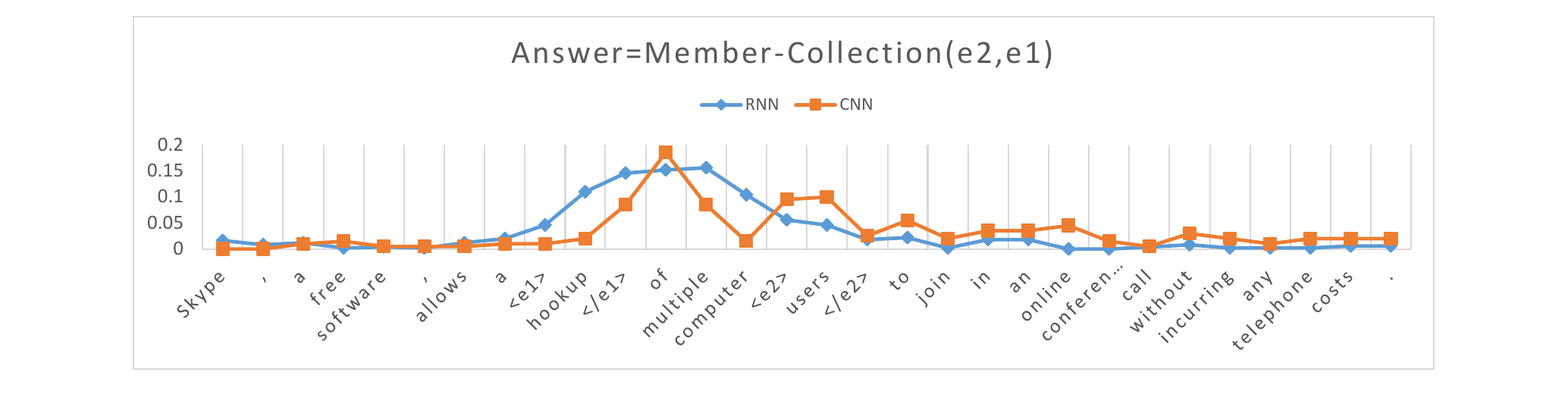}
\caption{Semantic distribution on words in the sentence ``Skype, a free software, allows
a $<$e1$>$ hookup $<$/e1$>$ of multiple computer $<$e2$>$ users $<$/e2$>$ to join in an
online conference call without incurring any telephone costs."  }
\label{fig:sentence2}
\end{figure}

To measure the contribution of a particular word or segment to the sentence-level
semantic meaning, for each sentence, we count the number of dimensions that the local feature at each
word step contributes to the output of the max-pooling. This number is divided
by the number of total dimensions
of the feature vector, resulting in a `semantic contribution' over the word sequence.
Figure~\ref{fig:sentence1} and Figure~\ref{fig:sentence2} show two examples
of semantic contributions. In each figure, the results with both the CNN and RNN
models are presented.

For the sentence in Figure~\ref{fig:sentence1}, the correct relation is `Instrument-Agency'. But CNN gives wrong answer `Other'.
It can be seen that CNN matches two patterns `is able to' and `magic', while the RNN matches
the entire sequence between the two nominals {\bf witch} and {\bf magic}, with the peak
at `by using'. Clearly, the pattern that the RNN model matches is more reasonable than that
matched by the CNN model.

We highlight that RNN is a temporal model which accumulates the semantic meanings word by word, so
the peak at `by using' is actually the contribution of all the words after `witch'. In
contrast, CNN model learns only local patterns, therefore it splits the semantic meaning into
two separate word segments.

Similar observation is obtained with second example shown in Figure~\ref{fig:sentence2}. Again,
the RNN model accumulates the semantic meaning of the sentence word by word, while the
CNN model has to learn two local patterns and merge them together.

An interesting observation is that the RNN-based semantic distribution tends to be
smoother than the one produced by the CNN model. In fact, we calculated the average
variance on the semantic contribution of neighbouring words
with all the sentences in the SemEval-2010 task 8 dataset, and found that the
variance with the RNN model is $0.0017$, while this number is $0.0025$ with the CNN model.
The smoother semantic distribution is certainly due to the temporal nature of the RNN model.

\section{Conclusion}
\label{sect:pdf}

In this paper, we proposed a simple RNN-based approach for relation classification. Compared to other deep learning
models such as CNN, the RNN model can deal with long-distance patterns and so is particular suitable for
learning relations within a long context. Several important modifications were proposed to improve
the basic model, including a max-pooling feature aggregation, a position indicator approach to
specify target nominals, and a bi-directional architecture to learn both the forward and backward contexts.

Experimental results on two different datasets demonstrated that the RNN-based approach can achieve
better results than CNN-based approach, and for sentences with long-distance relations,
the RNN model exhibits clear advantages.

%Despite the promising results, we found that some relations are easy to get confused with each other, and the
%error rate of the `Other' relation is quite high. This is possibly caused by the simple
%logistic regression classifier used in this study, which should be addressed in the future. Another potential research is to
%combine the CNN and RNN models, which allows leveraging the advantage of the two models in feature learning and temporal
%modelling respectively.

\newpage
\bibliographystyle{acl}
\bibliography{reference}

\end{document}